# Causal Transportability of Experiments on Controllable Subsets of Variables: $z$-Transportability

**Sanghack Lee and Vasant Honavar**
Artificial Intelligence Research Laboratory
Department of Computer Science
Iowa State University
Ames, IA 50011

## Abstract

We introduce *z-transportability*, the problem of estimating the causal effect of a set of variables $\mathbf{X}$ on another set of variables $\mathbf{Y}$ in a *target domain* from experiments on any subset of controllable variables $\mathbf{Z}$ where $\mathbf{Z}$ is an *arbitrary* subset of observable variables $\mathbf{V}$ in a *source domain*. $z$-Transportability generalizes $z$-identifiability, the problem of estimating in a given domain the causal effect of $\mathbf{X}$ on $\mathbf{Y}$ from surrogate experiments on a set of variables $\mathbf{Z}$ such that $\mathbf{Z}$ is disjoint from $\mathbf{X}$. $z$-Transportability also generalizes transportability which requires that the causal effect of $\mathbf{X}$ on $\mathbf{Y}$ in the target domain be estimable from experiments on any subset of *all* observable variables in the source domain. We first generalize $z$-identifiability to allow cases where $\mathbf{Z}$ is not necessarily disjoint from $\mathbf{X}$. Then, we establish a necessary and sufficient condition for $z$-transportability in terms of generalized $z$-identifiability and transportability. We provide a sound and complete algorithm that determines whether a causal effect is *z-transportable*; and if it is, produces a transport formula, that is, a recipe for estimating the causal effect of $\mathbf{X}$ on $\mathbf{Y}$ in the target domain using information elicited from the results of experimental manipulations of $\mathbf{Z}$ in the source domain and observational data from the target domain. Our results also show that *do*-calculus is complete for $z$-transportability.

## 1 INTRODUCTION

Elicitation of a causal effect from observations and experiments is central to scientific discovery, or more generally, rational approaches to understanding and interacting with the world around us. *Causal diagrams* (Pearl, 1995, 2000) provide a formal representation for combining data with causal information. *Do*-calculus (Pearl, 1995, 2000, 2012) provides a sound (Pearl, 1995) and *complete* (Shpitser and Pearl, 2006b; Huang and Valtorta, 2006) inferential machinery for causal inference. The resulting framework has been used to estimate *causal effects* of a set of variables $\mathbf{X}$ on another set of variables $\mathbf{Y}$ from *observations* and *interventions* (Pearl, 2000; Tian and Pearl, 2002; Tian, 2004; Shpitser and Pearl, 2006a).

In real world scenarios in which the treatment variables $\mathbf{X}$ may not be amenable to interventions due to technical or ethical considerations, it is interesting to consider experiments on a possibly different set of variables $\mathbf{Z}$ that are more amenable to manipulate than the treatment variables $\mathbf{X}$. Bareinboim and Pearl (2012a) introduced *z-identifiability,* the problem of estimating in a given domain (setting, environment, population) the causal effect of $\mathbf{X}$ on $\mathbf{Y}$ from surrogate experiments on $\mathbf{Z}$. In scenarios in which causal information acquired from one domain might be useful another different but related domain. Pearl and Bareinboim (2011) introduced *selection diagrams* for expressing knowledge about differences and commonalities between a source and a target domain. They used the selection diagrams to provide a formal definition of *transportability*, a license to transport causal information elicited from *experimental* studies in a *source* to a *target* domain in which only an *observational* study is possible. They also provided an algorithm for determining whether a causal effect is transportable given a selection diagram that represents a set of assumptions about the differences between the source and the target domains; and if so, computing a *transport formula,* which provides a recipe for estimating a causal effect in the target domain. In transporting the causal effect of a set of variables $\mathbf{X}$ on another set of variables $\mathbf{Y}$ from a source to a target domain we are free to use information acquired from *all possible* experiments on any subset of $\mathbf{V}$ in the source domain, available knowl-edge about differences and commonalities between the source and target domains (encoded by the selection diagram), and observations in both domains. However, many scenarios of practical interest present the problem of estimating in a target domain the causal effect of a set of variables $\mathbf{X}$ on another set of variables $\mathbf{Y}$ using experiments in the source domain on a subset $\mathbf{Z}$ of $\mathbf{V}$.

Against this background, we introduce *z-transportability*, the problem of estimating in a *target domain* the causal effect of a set of variables $\mathbf{X}$ on another set of variables $\mathbf{Y}$ from experiments on an arbitrary set of controllable variables $\mathbf{Z}$ in a *source domain*. $z$-Transportability generalizes *z-identifiability* (Bareinboim and Pearl, 2012a), the problem of estimating in a given domain, the causal effect of $\mathbf{X}$ on $\mathbf{Y}$ from surrogate experiments on $\mathbf{Z}$. $z$-Transportability also generalizes transportability (Pearl and Bareinboim, 2011) which requires only that the causal effect of $\mathbf{X}$ on $\mathbf{Y}$ in the target domain be estimable from experiments on $\mathbf{V}$ (where $\mathbf{V}$ is the set of *all* variables, including those included in $\mathbf{X}$ and those included in $\mathbf{Z}$) in the source domain. We first generalize $z$-identifiability to allow cases where $\mathbf{Z}$ is not necessarily disjoint from $\mathbf{X}$. Then, we establish a necessary and sufficient condition for $z$-transportability and relate it to the corresponding conditions for generalized $z$-identifiability (Bareinboim and Pearl, 2012a) and for transportability (Bareinboim and Pearl, 2012b). We provide a correct and *complete* algorithm that determines whether a causal effect is *z-transportable*; and if it is, produces a *transport formula*, that is, a recipe for estimating the causal effect of $\mathbf{X}$ on $\mathbf{Y}$ in the target domain using information elicited from the results of experimental manipulations of $\mathbf{Z}$ in the source domain and observational data from the target domain.

This work was carried out independently of Bareinboim and Pearl (2013a).[1] The key differences between (Bareinboim and Pearl, 2013a) and this paper are that (i) we establish a necessary and sufficient condition for $z$-transportability directly from existing results for generalized $z$-identifiability and transportability whereas Bareinboim and Pearl (2013a) introduces a graphical criterion called *zs*-hedge. In addition, our algorithm differs from that described in (Bareinboim and Pearl, 2013a) in how it goes about determining whether a causal effect is *z-transportable* (and if it is, computing a transport formula).

---

[1] Our work was completed in January 2013 and submitted to UAI 2013 on March 1, 2013. We learned of the results in (Bareinboim and Pearl, 2013a) when it appeared as a Technical Report in April, 2013 after its acceptance for publication in AAAI 2013 while our UAI 2013 submission was still under review.

The rest of the paper is organized as follows: Section 2 reviews some of the basic notions, essential definitions, and basic results that set the stage for the rest of the paper; Section 3 generalizes $z$-identifiability to remove disjoint assumptions on $\mathbf{Z}$; Section 4 introduces $z$-transportability and establishes a set of necessary and sufficient conditions for $z$-transportability. Section 5 describes an algorithm for $z$-transportability and proves its soundness and completeness. Section 6 concludes with a summary and an outline of some promising directions for further research.

## 2 PRELIMINARIES

Here we introduce some basic notations, review some basic notions, essential definitions, and basic results that set the stage for the rest of the paper.

We adopt notational convention established in the literature on identifiability (Tian and Pearl, 2002; Shpitser and Pearl, 2006b; Bareinboim and Pearl, 2012b,a). Variables are denoted by capital letters, e.g., $X, Y$ and their valuations or realizations by their lowercase counterparts, e.g., $x, y$. Bold letters e.g., $\mathbf{X}$ are used to denote sets of variables and their values e.g., $\mathbf{x}$. A directed acyclic graph (DAG) is denoted by $G$ and its vertices are denoted by $\mathbf{V}$. The set of *ancestors* of a variable $W$ in a graph $G$ (including $W$) is denoted by $An(W)_G$. We use $An(\mathbf{W})_G$ to denote $\bigcup_{W \in \mathbf{W}} An(W)_G$. We denote by $G[\mathbf{Y}]$, a subgraph of $G$ containing nodes in $\mathbf{Y}$ and all edges between the corresponding nodes in $G$. Following (Pearl, 2000), we denote by $G_{\overline{\mathbf{X}}}$, the edge subgraph of $G$ where all incoming arrows into nodes in $\mathbf{X}$ are deleted; by $G_{\underline{\mathbf{Y}}}$, the edge subgraph of $G$ where all outgoing arrows from nodes in $\mathbf{Y}$ are deleted; and by $G_{\overline{\mathbf{X}}\underline{\mathbf{Y}}}$, the edge subgraph of $G$ where all incoming arrows into nodes in $\mathbf{X}$ and all outgoing arrows from nodes in $\mathbf{Y}$ are deleted.

We now proceed to review some key definitions and results.

A *causal diagram* (Pearl, 2000) $G$ is a semi-Markovian graph (i.e., a graph with directed as well as bidirected edges that does not have directed cycles) which encodes a set of causal assumptions. A *causal model* (Pearl, 2000) is a tuple $\langle \mathbf{U}, \mathbf{V}, F \rangle$ where $\mathbf{U}$ is a set of *background* or *hidden* variables that cannot be observed or experimented on but which can influence the rest of the model; $\mathbf{V}$ is a set of observed variables $\{V_1, \ldots V_n\}$ that are determined by variables in the model, i.e., variables in $\mathbf{U} \cup \mathbf{V}$; $F$ is a set of deterministic functions $\{f_1, \ldots, f_n\}$ where each $f_i$ specifies the value of the observed variable $V_i$ given the values of observable parents of $V_i$ and the values of hidden causes of $V_i$. A *probabilistic causal model* (Pearl, 2000) (PCM) is a tuple $M = \langle \mathbf{U}, \mathbf{V}, F, P(\mathbf{U}) \rangle$ where $P(\mathbf{U})$ is a joint distribution over $\mathbf{U}$.

*Intervention* (Pearl, 2000) on a set of variables $\mathbf{X} \subseteq \mathbf{V}$ of PCM $M = \langle \mathbf{U}, \mathbf{V}, F, P(\mathbf{U}) \rangle$ involves setting to $\mathbf{X} = \mathbf{x}$ and is denoted by *do*-operation $do(\mathbf{X} = \mathbf{x})$ or simply $do(\mathbf{x})$. A *causal effect* of $\mathbf{X}$ on a disjoint set of variables $\mathbf{Y} \subseteq \mathbf{V} \setminus \mathbf{X}$ is written as $P(\mathbf{y}|do(\mathbf{x}))$ or simply $P_{\mathbf{x}}(\mathbf{y})$. Intervention on a set of variables $\mathbf{X} \subseteq \mathbf{V}$ creates a *submodel* (Pearl, 2000) $M_{\mathbf{x}}$ of $M$ defined as follows: $M_{\mathbf{x}} = \langle \mathbf{U}, \mathbf{V}, F_{\mathbf{x}}, P(\mathbf{U}) \rangle$ where $F_{\mathbf{x}}$ is obtained by taking a set of distinct copies of functions in $F$ and replacing the functions that determine the value of variables in $\mathbf{X}$ by constant functions setting the variables to values $\mathbf{x}$. It is easy to see that a causal diagram $G$ that encodes the causal assumptions of model $M$ is modified to $G_{\overline{\mathbf{X}}}$ by intervention on $\mathbf{X}$.

**Definition 1** (Causal Effects Identifiability (Pearl, 2000)). Let $\mathbf{X}$, $\mathbf{Y}$ be two sets of disjoint variables, and let $G$ be the causal diagram. The causal effect of an action $do(\mathbf{X} = \mathbf{x})$ on a set of variables $\mathbf{Y}$ is said to be *identifiable* from $P$ in $G$ if $P_{\mathbf{x}}(\mathbf{y})$ is (uniquely) computable from $P(\mathbf{V})$ in any model that induces $G$.

*Do*-calculus (Pearl, 1995) offers a sound and complete (Shpitser and Pearl, 2006b; Huang and Valtorta, 2006) inferential machinery for deciding identifiability (Tian and Pearl, 2002; Tian, 2004; Shpitser and Pearl, 2006a) in the sense that, if a causal effect is identifiable, there exists a sequence of applications of the rules of *do*-calculus that transforms the causal effect into a formula that includes only observational quantities. Let $G$ be a causal diagram and $P$ be a distribution on $G$. Let $\mathbf{W}$, $\mathbf{X}$, $\mathbf{Y}$, and $\mathbf{T}$ be disjoint sets of variables in $G$. Then, the three rules of *do*-calculus are (Pearl, 1995):

(Rule 1) Insertion/deletion of observations:
$P_{\mathbf{x}}(\mathbf{y} \mid \mathbf{t}, \mathbf{w}) = P_{\mathbf{x}}(\mathbf{y} \mid \mathbf{w})$ if $(\mathbf{Y} \perp\!\!\!\perp \mathbf{T} \mid \mathbf{X}, \mathbf{W})_{G_{\overline{\mathbf{X}}}}$

(Rule 2) Intervention/observation exchange:
$P_{\mathbf{x},\mathbf{t}}(\mathbf{y} \mid \mathbf{w}) = P_{\mathbf{x}}(\mathbf{y} \mid \mathbf{t}, \mathbf{w})$ if $(\mathbf{Y} \perp\!\!\!\perp \mathbf{T} \mid \mathbf{X}, \mathbf{W})_{G_{\overline{\mathbf{X}}\underline{\mathbf{T}}}}$

(Rule 3) Insertion/deletion of interventions:
$P_{\mathbf{x},\mathbf{t}}(\mathbf{y} \mid \mathbf{w}) = P_{\mathbf{x}}(\mathbf{y} \mid \mathbf{w})$ if $(\mathbf{Y} \perp\!\!\!\perp \mathbf{T} \mid \mathbf{X}, \mathbf{W})_{G_{\overline{\mathbf{X}}, \overline{\mathbf{T}(\mathbf{W})}}}$
where $\mathbf{T}(\mathbf{W})$ represents $\mathbf{T} \setminus An(\mathbf{W})_{G_{\overline{\mathbf{X}}}}$.

Shpitser and Pearl (2006b) devised an efficient and complete algorithm, **ID**, for identifying causal effects. **ID** employs *c-component decomposition* of a graph and the resulting factorization of a causal effect (Tian and Pearl, 2002) which can be expressed in terms of standard probability manipulations and *do*-calculus.

**Definition 2** (*C*-component). Let $G$ be a semi-Markovian graph such that a subset of its bidirected arcs forms a spanning tree over all vertices in $G$. Then $G$ is a *c-component* (confounded component).

We denote the set of c-components in $G$ by $\mathcal{C}(G)$.

Pearl and Bareinboim (2011) defined *transportability* which offers a license to transport causal information learned from *experimental* studies in a *source* domain to a *target* domain in which only an *observational* study is possible. They also introduced a *selection diagram*, a graphical representation for combining a causal diagram in a source with a causal diagram in a target domain.

**Definition 3** (Selection Diagram (Pearl and Bareinboim, 2011)). Let $\langle M, M^* \rangle$ be a pair of structural causal models relative to domains $\langle \Pi, \Pi^* \rangle$, sharing a causal diagram $G$. $\langle M, M^* \rangle$ is said to induce a *selection diagram* $D$ if $D$ is constructed as follows: (i) every edge in $G$ is also an edge in $D$; (ii) $D$ contains an extra edge $S_i \to V_i$ whenever there might exist a discrepancy $f_i \neq f_i^*$ or $P(U^i) \neq P^*(U^i)$ between $M$ and $M^*$.

We call the set of such $S_i$ selection variables and denote them by $\mathbf{S}$.

**Definition 4** (Causal Effects Transportability (Pearl and Bareinboim, 2011)). Let $D$ be a selection diagram relative to domains $\langle \Pi, \Pi^* \rangle$. Let $\langle P, I \rangle$ be the pair of observational and interventional distributions of $\Pi$, and $P^*$ be the observational distribution of $\Pi^*$. The causal effect $R = P_{\mathbf{x}}(\mathbf{y})$ is said to be *transportable* from $\Pi$ to $\Pi^*$ in $D$ if $P^*_{\mathbf{x}}(\mathbf{y})$ is uniquely computable from $P$, $P^*$, $I$ in any model that induces $D$.

Bareinboim and Pearl (2012b) provided **sID**, an algorithm for transporting causal effects from one domain to another. **sID** is an extension of **ID** (Shpitser and Pearl, 2006b), an algorithm for identifying causal effects from experiments and observations.

Bareinboim and Pearl (2012a) introduced *z-identifiability*, the problem of estimating in a given domain the effect on a set of variables $\mathbf{Y}$ of interventions on a set of variables $\mathbf{X}$ from surrogate experiments on a different set, $\mathbf{Z}$, that is more accessible to manipulation than $\mathbf{X}$.

**Definition 5** (Causal Effects z-Identifiability (Bareinboim and Pearl, 2012a)). Let $\mathbf{X}$, $\mathbf{Y}$, and $\mathbf{Z}$ be disjoint subsets of observable variables $\mathbf{V}$, and let $G$ be the causal diagram. The causal effect of an action $do(\mathbf{X} = \mathbf{x})$ on a set of variables $\mathbf{Y}$ is said to be *z-identifiable* from $P$ in $G$ if $P_{\mathbf{x}}(\mathbf{y})$ is (uniquely) computable from $P(\mathbf{V})$ together with the set of interventional distributions $I_{\mathbf{Z}} = \{P(\mathbf{V} \setminus \mathbf{Z}' \mid do(\mathbf{Z}'))\}_{\mathbf{Z}' \in \mathcal{P}(\mathbf{Z}) \setminus \{\emptyset\}}$, in any model that induces $G$.

Bareinboim and Pearl (2012a) established a graphical necessary and sufficient condition for *z-identifiability* for arbitrary disjoint sets of variables $\mathbf{X}$, $\mathbf{Y}$, and $\mathbf{Z}$:

**Theorem 1.** *The causal effect* $R = P(\mathbf{y} \mid do(\mathbf{x}))$ *is*

$zI\mathcal{D}$ in $G$ if and only if one of the following conditions hold:

1. $R$ is identifiable in $G$; or

2. There exists $\mathbf{Z}' \subseteq \mathbf{Z}$ such that the following conditions hold, (a) $\mathbf{X}$ intercepts all directed paths from $\mathbf{Z}'$ to $\mathbf{Y}$, and (b) $R$ is identifiable in $G_{\overline{\mathbf{Z}'}}$.

Bareinboim and Pearl (2012a) also established the completeness of *do*-calculus relative to $z$-identifiability. They also provided $\mathbf{ID^z}$, a complete algorithm for computing the causal effect of $\mathbf{X}$ on $\mathbf{Y}$ using information provided by experiments on $\mathbf{Z}$ under the assumption that $\mathbf{Z} \cap \mathbf{X} = \emptyset$.

## 3 GENERALIZED $z$-IDENTIFIABILITY

We proceed to generalize $z$-identifiability to allow cases where $\mathbf{Z}$ is not necessarily disjoint from $\mathbf{X}$.[2]

**Definition 6** (Generalized Causal Effects $z$-Identifiability). Let $\mathbf{X}$, $\mathbf{Y}$, and $\mathbf{Z}$ be arbitrary subsets of observable variables $\mathbf{V}$ with $\mathbf{X} \cap \mathbf{Y} = \emptyset$, and let $G$ be the causal diagram. The causal effect of an action $do(\mathbf{X} = \mathbf{x})$ on a set of variables $\mathbf{Y}$ is said to be *gz-identifiable* from $P$ in $G$ if $P_\mathbf{x}(\mathbf{y})$ is (uniquely) computable from $P(\mathbf{V})$ together with the set of interventional distributions $I_\mathbf{Z} = \{P(\mathbf{V} \setminus \mathbf{Z}' \mid do(\mathbf{Z}'))\}_{\mathbf{Z}' \in \mathcal{P}(\mathbf{Z}) \setminus \{\emptyset, \mathbf{V}\}}$, in any model that induces $G$.

We note that the assumption of $\mathbf{Y}$ and $\mathbf{Z}$ being disjoint can be trivially ignored since experiments on $\mathbf{Y} \cap \mathbf{Z}$ have no bearing on identification of a causal effect on $\mathbf{Y}$. In addition, the assumption in the definition of $z$-identifiability that $\mathbf{Z}$ and $\mathbf{X}$ are disjoint can be *easily* dropped since identifying $P_\mathbf{x}(\mathbf{y})$ in $G$ is *identical* to identifying $P_{\mathbf{x} \setminus \mathbf{z}}(\mathbf{y})$ in $G_{\overline{\mathbf{Z} \cap \mathbf{X}}}$.

The necessary and sufficient conditions for *gzID* can follow immediately from Theorem 1 with minor modifications to allow for the possibility that $\mathbf{Z}$ may not necessarily be disjoint from $\mathbf{X}$:

**Theorem 2.** *The causal effect $R = P(\mathbf{y} \mid do(\mathbf{x}))$ is gzID in $G$ if and only if one of the following conditions hold:*

1. *$R$ is identifiable in $G$; or*

---

[2]We will use the abbreviations *ID*, *zID*, *gzID*, *TR*, and *zTR* respectively to denote identifiability, $z$-identifiability, $gz$-identifiability, transportability, and $z$-transportability, respectively, when used as nouns; and identifiable, $z$-identifiable, $gz$-identifiable, transportable, $z$-transportable, respectively, when used as adjectives.

---

**function $\mathbf{GID^z}$** ($\mathbf{y}$, $\mathbf{x}$, $\mathbf{Z}$, $\mathcal{I}$, $\mathcal{J}$ $P$, $G$)
INPUT: $\mathbf{x}$,$\mathbf{y}$: value assignments; $\mathbf{Z}$: variables with interventions available; $\mathcal{I}$, $\mathcal{J}$: active experiments; $P$: current probability distribution $do(\mathcal{I}, \mathcal{J}, \mathbf{x})$ (observational when $\mathcal{I} = \mathcal{J} = \emptyset$); $G$: a causal graph;
OUTPUT: Expression for $P_\mathbf{x}(\mathbf{y})$ in terms of $P$, $P_\mathbf{z}$ or **FAIL**$(F, F')$.
1 **if** $\mathbf{X} = \emptyset$, **return** $\sum_{\mathbf{v} \setminus \mathbf{y}} P(\mathbf{v})$
2 **if** $\mathbf{V} \setminus An(\mathbf{Y})_G \neq \emptyset$,
  **return** $\mathbf{GID^z}(\mathbf{y}, \mathbf{x} \cap An(\mathbf{Y})_G, \mathbf{Z},$
    $\mathcal{I}, \mathcal{J}, \sum_{\mathbf{v} \setminus An(\mathbf{Y})_G} P, An(\mathbf{Y})_G)$
3 Set $\mathbf{W} = \mathbf{V} \setminus (\mathbf{X} \cup \mathcal{I} \cup \mathcal{J})) \setminus An(\mathbf{Y})_{G_{\overline{\mathbf{X} \cup \mathcal{I} \cup \mathcal{J}}}}$
  Set $\mathbf{Z_w} = \mathbf{Z} \cap (\mathbf{X} \cup \mathbf{W})$
  **if** $(\mathbf{Z_w} \cup \mathbf{W}) \neq \emptyset$,
  **return** $\mathbf{GID^z}(\mathbf{y}, \mathbf{x} \cup \mathbf{w} \setminus \mathbf{Z_w}, \mathbf{Z} \setminus \mathbf{Z_w},$
    $\mathcal{I} \cup \mathbf{z_w}, \mathcal{J}, P, G)$
4 **if** $\mathcal{C}(G \setminus (\mathbf{X} \cup \mathcal{I} \cup \mathcal{J})) = \{C_0, \ldots, C_k\}$,
  **return** $\sum_{\mathbf{v} \setminus \{\mathbf{y}, \mathbf{x}, \mathcal{I}\}} \prod_i \mathbf{GID^z}(c_i, (\mathbf{v} \setminus c_i) \setminus \mathbf{Z},$
    $\mathbf{Z} \setminus (\mathbf{V} \setminus C_i), \mathcal{I}, \mathcal{J} \cup (\mathbf{Z} \cap (\mathbf{v} \setminus c_i)), P, G)$
  **if** $\mathcal{C}(G \setminus (\mathbf{X} \cup \mathcal{I} \cup \mathcal{J})) = \{C\}$,
5 **if** $\mathcal{C}(G) = \{G\}$, **throw FAIL**$(G, C)$
6 **if** $C \in \mathcal{C}(G)$,
  **return** $\sum_{c \setminus \mathbf{y}} \prod_{i | V_i \in C} P(v_i \mid v_G^{(i-1)} \setminus (\mathcal{I} \cup \mathcal{J}))$
7 **if** $(\exists C') C \subset C' \in \mathcal{C}(G)$,
  **return** $\mathbf{GID^z}(\mathbf{y}, \mathbf{x} \cap C', \mathbf{Z}, \mathcal{I}, \mathcal{J},$
    $\prod_{i | V_i \in C'} P(V_i \mid V_G^{(i-1)} \cap C', v_G^{(i-1)} \setminus (C' \cup \mathcal{I} \cup \mathcal{J})), C')$

Figure 1: $\mathbf{GID^z}$ for *gzID*.

2. *There exists $\mathbf{Z}' \subseteq \mathbf{Z}$ such that the following conditions hold, (a) $\mathbf{X}$ intercepts all directed path from $\mathbf{Z}' \setminus \mathbf{X}$ to $\mathbf{Y}$, and (b) $P(y \mid do(\mathbf{x} \setminus \mathbf{z}'))$ is identifiable in $G_{\overline{\mathbf{Z}'}}$*

One may simply call $\mathbf{ID^z}$ by passing $P_{\mathbf{x} \setminus \mathbf{z}}(\mathbf{y})$ in $G_{\overline{\mathbf{Z} \cap \mathbf{X}}}$ with surrogate variables $\mathbf{Z} \setminus \mathbf{X}$ (i.e., wrapping $\mathbf{ID^z}$) to yield a sound and complete algorithm for *gzID*. Instead, we obtain $\mathbf{GID^z}$ (Figure 1) by making a minor modification to the $\mathbf{ID^z}$ algorithm (Bareinboim and Pearl, 2012a) (on line 3) which reflects Theorem 2. This only delays the use of experiments on $\mathbf{X} \cap \mathbf{Z}$ to line 3 to allow for the possibility that $\mathbf{Z} \cap \mathbf{X} \neq \emptyset$.

## 4 $z$-TRANSPORTABILITY

We introduce *z-transportability*, the problem of estimating the effect on a set of variables $\mathbf{Y}$ of interventions on a set of variables $\mathbf{X}$ in a *target domain* from experiments on an arbitrary set of controllable variables $\mathbf{Z}$ in a *source domain*.

**Definition 7** (Causal Effects $z$-Transportability). Let $\mathbf{X}$, $\mathbf{Y}$, $\mathbf{Z}$ be sets of variables where $\mathbf{X}$ is disjoint from $\mathbf{Y}$. Let $D$ be a selection diagram relative to domains $\langle \Pi, \Pi^* \rangle$. Let $P$ be the observational distribution and $I_\mathbf{Z} = \{P(\mathbf{V} \setminus \mathbf{Z}' \mid do(\mathbf{Z}'))\}_{\mathbf{Z}' \in \mathcal{P}(\mathbf{Z}) \setminus \{\emptyset, \mathbf{V}\}}$ the set of interventional distributions of $\Pi$, and $P^*$ the observational distribution of $\Pi^*$. The causal effect $P_\mathbf{x}(\mathbf{y})$ is said to be *z-transportable* from $\Pi$ to $\Pi^*$ in $D$ if $P_\mathbf{x}^*(\mathbf{y})$ is uniquely computable from $P$, $P^*$, $I_\mathbf{Z}$ in any model that induces $D$.

Thus, $z$-transportability requires that the causal effect of $\mathbf{X}$ on $\mathbf{Y}$ in a target domain be estimable from experiments on $\mathbf{Z}$ in a source domain when only the variables $\mathbf{Z}$ are controllable and any subset of $\mathbf{Z}$ can be controlled together. It is easy to see that $z$-transportability generalizes *gzID*, the problem of estimating in a given domain, the causal effect of $\mathbf{X}$ on $\mathbf{Y}$ from experiments on $\mathbf{Z}$. $z$-Transportability also generalizes *TR* which requires only that the causal effect of $\mathbf{X}$ on $\mathbf{Y}$ in the target domain be estimable from experiments on $\mathbf{V}$ in the source domain.

**Lemma 1.** *Let $\mathbf{X}$, $\mathbf{Y}$, $\mathbf{Z}$ be sets of variables with $\mathbf{X}$ disjoint from $\mathbf{Y}$, in population $\Pi$ and $\Pi^*$, and let $D$ be the selection diagram characterizing $\Pi$ and $\Pi^*$. $P(\mathbf{y} \mid do(\mathbf{x}))$ is not z-transportable from $\Pi$ to $\Pi^*$ if there exist two causal models $M^1$ and $M^2$ compatible with $D$ such that $P_1^*(\mathbf{V}) = P_2^*(\mathbf{V})$, $P_1(\mathbf{V}) = P_2(\mathbf{V})$, $P_1(\mathbf{V} \setminus \mathbf{Z}' \mid do(\mathbf{Z}')) = P_2(\mathbf{V} \setminus \mathbf{Z}' \mid do(\mathbf{Z}'))$, for all $\mathbf{Z}' \subseteq \mathbf{Z}$ and $\mathbf{Z}' \neq \mathbf{V}$, and $P_1^*(\mathbf{y} \mid do(\mathbf{x})) \neq P_2^*(\mathbf{y} \mid do(\mathbf{x}))$.*

*Proof.* The non-uniqueness of $P^*(\mathbf{y} \mid do(\mathbf{x}))$ implies that there is no function that maps from $P$, $P^*$, $I_\mathbf{Z}$ to $P^*(\mathbf{y} \mid do(\mathbf{x}))$. □

**Lemma 2.** *Let $\mathbf{X}$, $\mathbf{Y}$, $\mathbf{Z}$ be sets of variables with $\mathbf{X}$ disjoint from $\mathbf{Y}$. Let $D$ be a selection diagram characterizing $\Pi$ and $\Pi^*$, and $\mathbf{S}$ be a set of selection variables in $D$. The causal effect $R = P(\mathbf{y} \mid do(\mathbf{x}))$ is z-transportable from $\Pi$ to $\Pi^*$ in $D$ if $P(\mathbf{y} \mid do(\mathbf{x}), \mathbf{s})$ is reducible, using the rules of do-calculus, to an expression in which: $\mathbf{S}$ appears only as a conditioning variable in do-free terms; and interventions in do-terms are a subset of $\mathbf{Z}$.*

*Proof.* An expression that is a transport formula for $P(\mathbf{y} \mid do(\mathbf{x}), \mathbf{s})$ can contain only $P^*$, $P$ (terms without the *do*-operator) and $I_\mathbf{Z}$ (terms that contain the *do*-operator on a subset of $\mathbf{Z}$ and but not the selection variables). By the correctness of *do*-calculus, the existence of the formula implies $z$-transportability of $P^*(\mathbf{y} \mid do(\mathbf{x}))$. □

Figure 2 shows selection diagrams where $P_x(y)$ is not *ID* but *zTR* given an experiment on $Z$. The causal effect $P_x^*(y)$ in each graph is uniquely estimable using the rules of *do*-calculus. By adding experiments on $Z$ by rule 3, we get: $P_x(y \mid s) = P_{z,x}(y \mid s)$. We can eliminate the effect of selection variable (■) on the two domains using rule 1 to obtain: $P_{z,x}(y \mid s) = P_{z,x}(y)$. Except in Figure 2(b), $P_{z,x}(y)$ can be expressed using rule 2 as: $P_{z,x}(y) = P_z(y \mid x)$. In Figure 2(b) we have: $P_{z,x}(y) = \sum_w P_z(w) P_z(y \mid w, x)$.

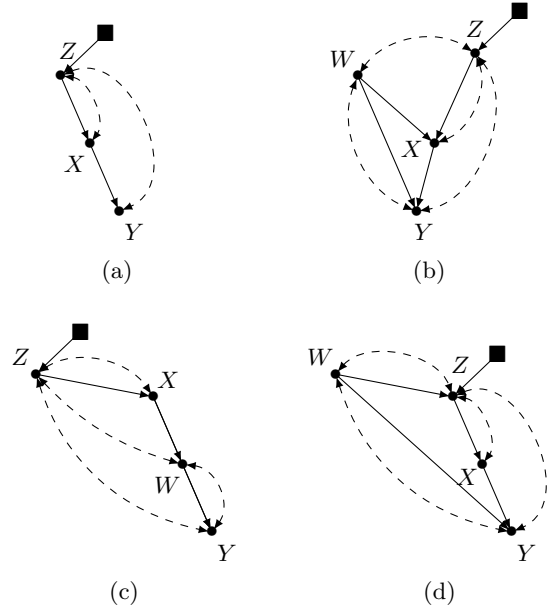

Figure 2: Selection diagrams where $P_x(y)$ is *zTR*, but not *ID*, with an experiment on $Z$. A selection variable $S$ is represented by a black square ■.

If $P_x(y)$ is *non-gzID* or *non-TR*, then the causal effect is *non-zTR* (see Lemma 3). For example, in the case of the four-node selection diagrams in Figure 2, if a controllable variable is $W$ instead of $Z$, then $P_x(y)$ is *non-gzID* and hence *non-zTR*. Similarly, if a selection variable is pointing to $W$ instead of $Z$, then $P_x(y)$ is *non-TR* and hence *non-zTR*.

We now proceed to establish the necessary and sufficient conditions for *zTR*. We start with a lemma that asserts a necessary condition for *zTR* in terms of *TR* and *gzID*.

**Lemma 3** (Necessity). *A causal effect $R = P_\mathbf{x}(\mathbf{y})$ is z-transportable from $\Pi$ to $\Pi^*$ in $D$ only if $R$ is gz-identifiable from $P$ and $I_\mathbf{Z}$ in $G$ and $R$ is transportable from $\Pi$ to $\Pi^*$ in $D$.*

*Proof.* This follows from the definitions of *gzID*, *TR* and *zTR*. First, *gzID* is a special case of *zTR* where $\Pi = \Pi^*$ ($\mathbf{S} = \emptyset$). In addition, *TR* is a special case of *zTR* where $\mathbf{Z} = \mathbf{V}$. Since no difference between two domains, $\mathbf{S} = \emptyset$, or availability of all experiments, $\mathbf{Z} = \mathbf{V}$, make the problem easier, *zTR* of $R$ implies *gzID* of $R$ and *TR* of $R$. It then follows that general $z$-identifiability of $R$ and transportability of $R$ are *necessary* for $z$-transportability of $R$. □

Since every *zTR* relation satisfies *gzID* and *TR*, we proceed to examine *TR* and *gzID* in depth. Bareinboim and Pearl (2012b) observed that a *TR* causal relation can be decomposed into *trivially transportable* and *directly transportable* (*DTR* for short) relations.

**Definition 8** (Trivial Transportability). A causal relation $R$ is said to be *trivially transportable* from $\Pi$ to $\Pi^*$, if $R(\Pi^*)$ is identifiable from $(G^*, P^*)$.

**Definition 9** (Direct Transportability). A causal relation $R$ is said to be *directly transportable* from $\Pi$ to $\Pi^*$, if $R(\Pi^*) = R(\Pi)$.

The equality of relations $P^*_\mathbf{x}(\mathbf{y})$ and $P_\mathbf{x}(\mathbf{y})$ holds if $(\mathbf{S} \perp\!\!\!\perp \mathbf{Y} \mid \mathbf{X})_{D_{\overline{\mathbf{X}}}}$ by rule 1 of *do*-calculus.

Recall that a causal effect $R$ can be factorized into multiple causal effects (*c-factors*) based on *c-components* (Tian and Pearl, 2002) and *gzID* of $R$ can be determined using a divide-and-conquer strategy. Hence, a causal effect is *gzID* if and only if each *c-factor* resulting from the factorization of $R$ is identifiable by the condition 1 or 2 in Theorem 2. It therefore follows that given a causal relation $R$ that is both *TR* and *gzID*, if one of the *c-factors* of $R$ is not *ID* in the target domain, then that *c-factor* must be *DTR* from the source domain and *ID* from $P_{\mathbf{Z}'}$ in an edge subgraph $G'_{\overline{\mathbf{Z}'}}$ of a causal diagram $G$ (condition 2 in Theorem 2). We provide a basic lemma for the factorization of a causal effect based on (Tian and Pearl, 2002; Shpitser and Pearl, 2006b).

**Lemma 4.** *Let $\mathbf{X}$ and $\mathbf{Y}$ be disjoint sets of variables of $G$. Let $G' = G[An(\mathbf{Y})_G]$, $\mathbf{X}' = \mathbf{X} \cap An(\mathbf{Y})_G$, and $\mathbf{V}'$ be variables in $G'$ and $\mathbf{v}'$ their valuations. Let $\mathbf{W} = \mathbf{V}' \setminus \mathbf{X}' \setminus An(\mathbf{Y})_{G'_{\overline{\mathbf{X}'}}}$, $\mathbf{X}'' = \mathbf{X}' \cup \mathbf{W}$; $\mathcal{C}(G' \setminus \mathbf{X}'')$ a c-component decomposition of graph $G' \setminus \mathbf{X}''$; and $\mathbf{Q} = \{P_{\mathbf{v}' \setminus c_i}(c_i)\}_{C_i \in \mathcal{C}(G' \setminus \mathbf{X}'')}$ the set of corresponding c-factors. Then $P_\mathbf{x}(\mathbf{y}) = \sum_{\mathbf{v}' \setminus \{\mathbf{x}'' \cup \mathbf{y}\}} \prod_{Q_i \in \mathbf{Q}} Q_i$.*

*Proof.* The proof follows from Lemma 3 (Shpitser and Pearl, 2006b). □

**Lemma 5.** *Let $G$ be a common causal diagram of domains $\Pi$ and $\Pi^*$. Let $\mathbf{X}$, $\mathbf{Y}$, $\mathbf{Z}$ be sets of variables of $G$ with $\mathbf{X}$ disjoint from $\mathbf{Y}$ and $\mathbf{v}$ a valuation of $\mathbf{V}$. Let $G' = G[An(\mathbf{Y})_G]$, $\mathbf{X}' = \mathbf{X} \cap An(\mathbf{Y})_G$, and $\mathbf{V}'$ be variables in $G'$ and $\mathbf{v}'$ their valuations. Let $\mathbf{W} = \mathbf{V}' \setminus \mathbf{X}' \setminus An(\mathbf{Y})_{G'_{\overline{\mathbf{X}'}}}$, $\mathbf{X}'' = \mathbf{X}' \cup \mathbf{W}$; $\mathcal{C}(G' \setminus \mathbf{X}'')$ a c-component decomposition of graph $G' \setminus \mathbf{X}''$; and $\mathbf{Q} = \{P_{\mathbf{v}' \setminus c_i}(c_i)\}_{C_i \in \mathcal{C}(G' \setminus \mathbf{X}'')}$ the set of corresponding c-factors. If $R = P_\mathbf{x}(\mathbf{y})$ is gzID from $\Pi$ and TR from $\Pi$ to $\Pi^*$, then for every $Q_i \in \mathbf{Q}$ the following conditions hold:*

(i) *$Q_i$ is identifiable from $P^*(\mathbf{V}')$ in $G'$; or*

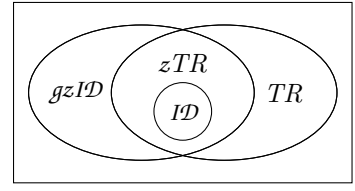

Figure 3: A Venn diagram depicting Theorem 3. Given a selection diagram $D$ and controllable variables $\mathbf{Z}$, the intersection of *gzID* and *TR* relations exactly matches to *zTR* relations in any model that induces $D$.

(ii) *$Q_i$ is (a) identifiable from $P_{\mathbf{z}'}$ in $G'_{\overline{\mathbf{Z}'}}$ where $\mathbf{Z}' = (\mathbf{V}' \setminus C_i) \cap \mathbf{Z}$ and (b) directly transportable from $\Pi$ to $\Pi^*$.*

*Proof.* Let $\mathbf{Q}_1$ be a set of *all ID* causal effects in $\mathbf{Q}$ and let $\mathbf{Q}_2$ be $\mathbf{Q} \setminus \mathbf{Q}_1$. If each causal effect in $\mathbf{Q}_2$ does not satisfy the second condition, it contradicts the premise that $R$ is *gzID* and *TR*. □

**Lemma 6** (Sufficiency). *Let $\mathbf{X}$, $\mathbf{Y}$, and $\mathbf{Z}$ be sets of variables of $G$ with $\mathbf{X}$ disjoint from $\mathbf{Y}$. If $R = P_\mathbf{x}(\mathbf{y})$ is gzID in $\Pi$ and TR from $\Pi$ to $\Pi^*$, then $R$ is zTR from $\Pi$ to $\Pi^*$.*

*Proof.* By Lemma 5, for every *c-factor* $Q \in \mathbf{Q}$ that is not *ID* in $\Pi^*$, there exists a subset $\mathbf{Z}$ of $\mathbf{V}$ such that $Q$ is identifiable in $\Pi$ using experiments on $\mathbf{Z}$ and *DTR* of $Q$. Hence, the causal effect $R$ is reducible to an expression where every term from *c-factors* in $\mathbf{Q}$ is either (i) a *do*-free term (i.e., identified from $P^*$) or (ii) a term that contains the *do*-operator but no selection variables (i.e., identified from $I_\mathbf{Z}$) in which the intervention is on a subset of $\mathbf{Z}$. Therefore, $R$ is *zTR* from $\Pi$ to $\Pi^*$. □

The necessary and sufficient conditions for *zTR* follow from (Lemma 3) and sufficiency (Lemma 6).

**Theorem 3** (Necessity and Sufficiency). *A causal effect $R = P_\mathbf{x}(\mathbf{y})$ is zTR from $\Pi$ to $\Pi^*$ in $D$ if and only if (i) $R$ is gzID from $P$ and $I_\mathbf{Z}$ in $G$ and (ii) $R$ is TR from $\Pi$ to $\Pi^*$ in $D$.*

Though Theorem 3 is the main theorem of the paper, it does not directly provide an effective procedure for estimating a causal effect given $P^*$, $P$, and $I_\mathbf{Z}$. Rather, Lemma 5 will be more instrumental in the design of a complete algorithm for *zTR*.

**function sID$^{\mathbf{z}}$ (y, x, $G$, Z)**

**Input**: y, x value assignments for a causal effect $P_{\mathbf{x}}^*(\mathbf{y})$; $G$ a causal diagram; Z a set of (inactive) controllable variables;
**Output**: an expression for $P_{\mathbf{x}}^*(\mathbf{y})$ regarding $P^*$ and $I_{\mathbf{Z}}$ or a (s-)hedge.

1  if $\mathbf{X} = \emptyset$, return $P^*(\mathbf{y})$
2  if $\mathbf{V} \neq An(\mathbf{Y})_G$, return sID$^{\mathbf{z}}$(y, $\mathbf{x} \cap An(\mathbf{Y})_G$, $G[An(\mathbf{Y})_G]$, Z)
3  $\mathbf{W} \leftarrow \mathbf{V} \setminus \mathbf{X} \setminus An(\mathbf{Y})_{G_{\overline{\mathbf{X}}}}$
4  if $\mathbf{W} \neq \emptyset$, return sID$^{\mathbf{z}}$(y, $\mathbf{x} \cup \mathbf{w}$, $G$, Z)
5  $\{C_1, \ldots, C_k\} \leftarrow \mathcal{C}(G \setminus \mathbf{X})$
6  $T_i \leftarrow \{V_j\}_{S_j \in \mathbf{S}} \cap C_i = \emptyset$ for each $i \in \{1, \ldots, k\}$
7  return $\sum_{\mathbf{v} \setminus \{\mathbf{y}, \mathbf{x}\}} \prod_{i=1}^{k} T_i ?$ **BI**$(c_i, \mathbf{v} \setminus c_i \setminus \mathbf{Z}, P_{\mathbf{z} \cap (\mathbf{v} \setminus c_i)}(\mathbf{V}), G_{\overline{\mathbf{Z} \cap (\mathbf{V} \setminus C_i)}}, \mathbf{Z} \cap (\mathbf{V} \setminus C_i)) :$ **BI**$(c_i, \mathbf{v} \setminus c_i, P^*(\mathbf{V}), G)$

**function BI (y, x, $P$, $G$, $\mathcal{I} = \emptyset$)**

**Input**: $P$: current distribution; $G$: current causal diagram; $\mathcal{I}$: active experiments with default value $\emptyset$.
**Output**: Expression for $P_{\mathbf{x}}(\mathbf{y})$ in terms of $P$ or throw a (s-)hedge depending on the existence of selection variables.

1  if $\mathbf{X} = \emptyset$, return $P(\mathbf{y})$
2  if $\mathbf{V} \neq An(\mathbf{Y})_G$, return **BI** (y, $\mathbf{x} \cap An(\mathbf{Y})_G$, $P(An(\mathbf{Y})_G)$, $G[An(\mathbf{Y})_G]$, $\mathcal{I} \cap An(\mathbf{Y})_G$)
3  $\{C\} \leftarrow \mathcal{C}(G \setminus (\mathbf{X} \cup \mathcal{I}))$
4  if $\mathcal{C}(G) = \{G\}$, throw **FAIL**$\langle D[G], D[C] \rangle$
5  if $C \in \mathcal{C}(G)$, return $\sum_{c \setminus \mathbf{y}} \prod_{i | V_i \in C} P(v_i \mid v_G^{(i-1)} \setminus \mathcal{I})$
6  if $(\exists C') C \subset C' \in \mathcal{C}(G)$, return **BI**(y, $\mathbf{x} \cap C'$, $\prod_{i | V_i \in C'} P(V_i \mid V_G^{(i-1)} \cap C', v_G^{(i-1)} \setminus (C' \cup \mathcal{I}))$, $C'$, $\mathcal{I} \cap C'$)

Figure 4: An algorithm for $zTR$ with a subroutine **BI** for the identification of a $c$-factor given a (interventional) probability distribution and a (mutilated) graph. To estimate a causal effect $P_{\mathbf{x}}^*(\mathbf{y})$, call **sID$^{\mathbf{z}}$** (y, x, $G$, Z). We assume that $P^*$, $I_{\mathbf{Z}}$, and $D$ are globally defined for convenience. By construction, $G = G_{\overline{\mathcal{I}}}$ in **BI**. Distribution $P_{\mathbf{z} \cap (\mathbf{v} \setminus c_i)}$ is obtained from $I_{\mathbf{Z}}$.

## 5 AN ALGORITHM FOR $z$-TRANSPORTABILITY

We proceed to describe **sID$^{\mathbf{z}}$**, an algorithm that determines whether a causal relation $P_{\mathbf{x}}(\mathbf{y})$ is $zTR$ from $\Pi$ to $\Pi^*$, and if so, produces a correct transport formula; if not, provides an evidence of *non-zTR* (i.e., an *s-hedge* (Bareinboim and Pearl, 2012b) or a *hedge* (Shpitser and Pearl, 2006b; Bareinboim and Pearl, 2012a) if the relation is not $TR$ or not $gzID$, respectively).

The design of **sID$^{\mathbf{z}}$** (see Figure 4[3,4]) follows directly from Lemma 5. Specifically, **sID$^{\mathbf{z}}$** factorizes a causal effect based on the decomposition of the given graph into a set of $c$-components. Unlike **GID$^{\mathbf{z}}$** (line 3, Figure 1), **sID$^{\mathbf{z}}$** (Figure 4) postpones covering interventions on $\mathbf{X}$ by experiments on $\mathbf{Z}$ until after the factorization of causal effect until it determines (line 7) whether each $c$-factor can in fact be identified from either the source domain or the target domain. From Lemma 5, each $c$-factor of a $z$-transportable causal effect should be a) identifiable (trivially transportable) in the target domain or b) $gz$-identifiable in the source domain and directly transportable from the source domain to the target domain. Fortunately, direct transportability of a $c$-factor $Q_i$ can be easily computed at the stage of decomposition:

$$\{V_j\}_{S_j \in \mathbf{S}} \cap C_i = \emptyset \qquad (1)$$

which is identical to testing S-admissibility (Pearl and Bareinboim, 2011)

$$(\mathbf{S} \perp\!\!\!\perp C_i \mid \mathbf{V} \setminus C_i)_{G_{\overline{\mathbf{V} \setminus C_i}}}.$$

From Theorem 3 above, to establish that a $c$-factor is not $zTR$, it suffices to show the existence of either 1) a hedge (which shows that the causal effect is non $gzID$); or 2) an s-hedge (which shows that the causal effect is non-$TR$). Algorithm **sID$^{\mathbf{z}}$** calls the subroutine **BI** to determine if a directly transportable $c$-factor is $gz$-identifiable. Because ordinarily identifiable $c$-factor is $gz$-identifiable, line 7 of **sID$^{\mathbf{z}}$** employs an inline conditional operator[5]. The subroutine **BI** estimates a $c$-factor given a distribution, a causal graph, and active experiments $\mathcal{I}$. Thus, the algorithm differs from **TR$^{\mathbf{z}}$** (Bareinboim and Pearl, 2013a) which tries to estimate a $c$-factor given an interventional distribution of a source domain *after* the test for trivial transportability of a $c$-factor fails.

The ability to check for direct transportability of a $c$-factor (Equation 1) allows **BI**, upon failure to iden-tify a c-factor in the source domain, to throw a hedge which implies non-$gzID$. Similarly, if a c-factor is neither directly-transportable from the source domain to the target domain nor ordinarily identifiable in the target domain, **BI** throws an s-hedge since $\{V_j\}_{S_j \in \mathbf{S}}$ intersects $C_i$.

---

[3]For simplicity, we combine $\mathcal{I}$ and $\mathcal{J}$ (see **GID$^{\mathbf{z}}$**, Figure 1) into $\mathcal{I}$ (see **sID$^{\mathbf{z}}$**, Figure 4)

[4]A manipulated graph $G_{\overline{\mathcal{I}}}$ and experimental distribution $P_{\mathcal{I}}$ are passed as arguments when *active* experiments are set to $\mathcal{I}$. For the use of $G_{\overline{\mathcal{I}}}$ ($G_{\overline{\mathcal{I} \cup \mathcal{J}}}$ in **GID$^{\mathbf{z}}$**) we must remove the incoming edges on the active experiments so that a relation can be identified from $P_{\mathbf{Z}'}$ in $G_{\overline{\mathbf{Z}'}}$.

[5]An inline conditional operator is of the form $cond?exp_1 : exp_2$. The first expression $exp_1$ is executed if $cond$ is true. Otherwise $exp_2$ is executed.

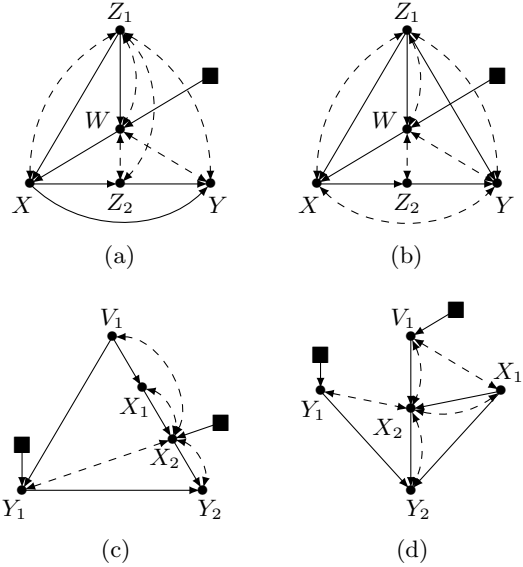

Figure 5: Selection diagrams where $P_{\mathbf{x}}(\mathbf{y})$ is not $\mathit{ID}$ but $zTR$ but given controllable variables $\mathbf{Z}$ ($V_1$ and $X_2$ are controllable variables for Figure 5(c) and 5(d)).

### 5.1 EXAMPLES

Some examples are given in Figure 5 to illustrate how $\mathbf{sID^z}$ estimates a causal effect $P_{\mathbf{x}}^*(\mathbf{y})$ from experiments on $\mathbf{Z}$ from a source domain. We will use $P_{(\mathcal{I}),\mathbf{x}\setminus\mathcal{I}}(\mathbf{y})$ to denote a causal effect $P_{\mathbf{x}}(\mathbf{y})$ in $G$ from an interventional distribution $P_{\mathcal{I}}$ in $G_{\overline{\mathcal{I}}}$ (such that $\mathcal{I} \subseteq \mathbf{X}$).

In Figure 5(a), $W$ and $Z_1$ are added as interventions (line 4), $P_x^*(y) = P_{z_1,w,x}^*(y)$. $Z_1$, which is included in controllable variables $\mathbf{Z}$, will not be treated as an active experiment until after the decomposition. The causal effect is factorized as $\sum_{z_2} P_{\mathbf{z},w,x}^*(y) P_{z_1,w,x,y}^*(z_2)$. Since no selection variable is pointing to $Y$ or $Z_2$ (line 6 and 7), the two c-factors are then identified from $P_{\mathbf{z}}$ in $G_{\overline{\mathbf{z}}}$ and $P_{z_1}$ in $G_{\overline{Z_1}}$, respectively, as

$$\sum_{z_2} P_{(z_1,z_2),w,x}(y) P_{(z_1),w,x,y}(z_2).$$

The parameters for the subroutine is $\mathbf{BI}(y, w \cup x, P_{\mathbf{z}}, G_{\overline{\mathbf{z}}}, \mathbf{z})$ and $\mathbf{BI}(z_2, w \cup x \cup y, P_{z_1}, G_{\overline{Z_1}}, z_1)$, respectively.

In Figure 5(b), $P_x^*(y) = P_{w,x}^*(y)$ by line 4. In lines 5–7, two c-factors $P_{z_2,w,x}^*(z_1, y)$ and $P_{z_1,w,x,y}^*(z_2)$ will be identified in the source domain as

$$\sum_{\mathbf{z}} P_{(z_2),w,x}(z_1, y) P_{(z_1),w,x,y}(z_2).$$

In Figure 5(c), $P_{\mathbf{x}}^*(\mathbf{y}) = P_{x_1,x_2}^*(y_1, y_2)$ is factorized as $\sum_{v_1} P_{\mathbf{x},\mathbf{y}}^*(v_1) P_{v_1,\mathbf{x},y_2}^*(y_1) P_{v_1,\mathbf{x},y_1}^*(y_2)$. Since a selection variable is pointing to $Y_1$,

$$P_{v_1,\mathbf{x},y_2}^*(y_1) \neq P_{v_1,\mathbf{x},y_2}(y_1) = P_{(v_1,x_2),x_1,y_2}(y_1).$$

Then, the first and third c-factors will be identified in the source domain as

$$\sum_{v_1} P_{(x_2),x_1,\mathbf{y}}(v_1) P_{v_1,\mathbf{x},y_2}^*(y_1) P_{(v_1,x_2),x_1,y_1}(y_2).$$

In Figure 5(d), $V_1$ is added to the causal effect as an intervention, $P_{\mathbf{x}}^*(\mathbf{y}) = P_{v_1,\mathbf{x}}^*(\mathbf{y})$ by line 4. By Lemma 4, $P_{v_1,\mathbf{x}}^*(\mathbf{y}) = P_{v_1,\mathbf{x},y_2}^*(y_1) P_{v_1,\mathbf{x},y_1}^*(y_2) = P_{v_1,\mathbf{x},y_2}^*(y_1) P_{(v_1,x_2),x_1,y_1}(y_2)$.

### 5.2 SOUNDNESS AND COMPLETENESS

We first illustrate a certain behavior of $\mathbf{GID^z}$ which will help understanding correctness of $\mathbf{sID^z}$.

*Remark* 1. Addition of interventions (line 3) and decomposition (line 4) of $\mathbf{GID^z}$ are executed at most once.

*Proof.* Given a causal relation, $\mathbf{GID^z}$ checks whether interventions can be added to the relation (line 3). If so, it adds interventions and the subsequent call checks whether there are multiple c-components in $G \setminus (\mathbf{X} \cup \mathcal{I} \cup \mathcal{J})$ (line 4). If so, the relation is factorized to c-factors; or the relation is already a c-factor. During the estimation of the c-factor, no interventions are added as $(\mathbf{X} \cup (\mathcal{I} \cup \mathcal{J})) \cup \mathbf{Y} = \mathbf{V}$. In addition, decomposition is not required as $G[\mathbf{Y}]$ is a c-component. □

**Lemma 7.** *Decomposition of $P_{\mathbf{x}}(\mathbf{y})$ produced by $\mathbf{sID^z}$ is equivalent to that produced by $\mathbf{sID}$ and by $\mathbf{GID^z}$.*

*Proof.* Trivially, the decomposition by $\mathbf{sID}$ and by $\mathbf{sID^z}$ are identical as $\mathbf{sID}$ and $\mathbf{sID^z}$ share the same code. As in Remark 1, lines 3 and 4 of $\mathbf{GID^z}$ are executed only once. Hence, when $\mathbf{GID^z}$ adds interventions, $\mathcal{I}$ and $\mathcal{J}$ are empty (line 3) and when it decomposes the relation, $\mathcal{J}$ is empty (line 4). Therefore, $\mathbf{X} \cup \mathbf{W}$ computed by $\mathbf{GID^z}$ (line 3) and by $\mathbf{sID^z}$ (line 4) are identical. Since $\mathcal{I}$ is set to $\mathbf{Z_w} \subseteq \mathbf{X} \cup \mathbf{W}$ by $\mathbf{GID^z}$, $\mathbf{X} \cup \mathcal{I}$ (line 4 of $\mathbf{GID^z}$) is identical to $\mathbf{X}$ in $\mathbf{sID^z}$ (line 5). As a result, $\mathcal{C}(G \setminus (\mathbf{X} \cup \mathcal{I} \cup \mathcal{J}))$ in line 4 of $\mathbf{GID^z}$ is equivalent to $\mathcal{C}(G \setminus \mathbf{X})$ in line 5 of $\mathbf{sID^z}$. □

**Lemma 8.** *The c-factor estimate produced by $\mathbf{BI}$ are equivalent to those produced by $\mathbf{GID^z}$.*

*Proof.* The code used by **BI** to estimate a *c*-factor is identical to that used by **GID$^z$** except for lines 3 and 4 of **GID$^z$** (see Footnote 3 and 4). By Remark 1 and Lemma 7, the *c*-factor estimate produced by **BI** is identical to that produced by **GID$^z$**. □

**Theorem 4** (Soundness). *Whenever* **sID$^z$** *returns an expression for a causal effect $P_\mathbf{x}^*(\mathbf{y})$, it is correct.*

*Proof.* The proof follows from Lemma 4 and 5, **sID$^z$** decomposes a causal effect $P_\mathbf{x}^*(\mathbf{y})$ and estimates its *c*-factors either from a source domain or from a target domain. If a *c*-factor is directly transportable, **sID$^z$** uses experimental distributions from the source domain to estimate it and the call to **BI** yields a *c*-factor estimate that is identical to that of **GID$^z$** (Lemma 8). If a *c*-factor is not directly transportable (Equation 1), the *c*-factor must be trivially identifiable in the target domain and since there are no active experiments, the *c*-factor estimate produced by **BI** is identical to that produced by **ID** (Lemma 8 with $\mathcal{I} = \emptyset$). □

**Theorem 5.** *Assume* **sID$^z$** *fails to z-transport $P_\mathbf{x}(\mathbf{y})$ from $\Pi$ to $\Pi^*$. Then $P_\mathbf{x}(\mathbf{y})$ is neither gz-identifiable from P in G nor transportable from $\Pi$ to $\Pi^*$ in D.*

*Proof.* If **sID$^z$** fails to z-transport a relation $P_\mathbf{x}(\mathbf{y})$ (in line 4 of the subroutine **BI**), it throws a hedge or an s-hedge. The failure is due to the existence of a hedge or an s-hedge associated with a *c*-factor, say $Q$. From Lemma 8 it follows that the failure of **sID$^z$** in the case of *empty* active experiments ($\mathcal{I} = \emptyset$) or *nonempty* active experiments ($\mathcal{I} \neq \emptyset$) respectively implies and *non-ID* or *non-gzID* of $Q$. From the test of direct transportability in line 6 of **sID$^z$** (Equation 1), *non-ID* of $Q$ implies non-direct-transportability of $Q$. This also implies that $P_\mathbf{x}(\mathbf{y})$ is not transportable since there exists a *c*-factor $Q$ that is neither trivially transportable (*ID*) nor direct-transportable. Therefore, whenever the algorithm fails, $P_\mathbf{x}(\mathbf{y})$ is neither *gzID* nor *TR*. □

**Corollary 1** (Completeness). **sID$^z$** *is complete.*

*Proof.* This follows from necessity and sufficiency theorem (Theorem 3) and Theorem 5. □

The completeness of **sID$^z$** proves that *do*-calculus and standard probability manipulations are sufficient for determining whether a causal effect is *z*-transportable.

## 6 SUMMARY AND DISCUSSION

We have introduced *z*-transportability, the problem of estimating in a target domain the causal effect of a set of variables **X** on another set of variables **Y** (such that $\mathbf{Y} \cap \mathbf{X} = \emptyset$) from experiments on any subset of an *arbitrary* controllable variables **Z** (such that $\mathbf{Z} \subseteq \mathbf{V}$) in a source domain. *z*-Transportability generalizes *z*-identifiability (Bareinboim and Pearl, 2012a), the problem of estimating in a given domain the causal effect of **X** on **Y** from surrogate experiments on **Z**. *z*-Transportability also generalizes transportability (Pearl and Bareinboim, 2011) which requires only that the causal effect of **X** on **Y** in the target domain be estimable from experiments on *all* variables in the source domain. We have generalized *z*-identifiability to allow cases where **Z** is not necessarily disjoint from **X**. We have established a necessary and sufficient condition for *z*-transportability in terms of generalized *z*-identifiability and transportability. We have provided **sID$^z$**, an algorithm that determines whether a causal effect is *z-transportable*; and if it is, produces a transport formula, that is, a recipe for estimating the causal effect of **X** on **Y** in the target domain using information elicited from the results of experimental manipulations of **Z** in the source domain and observational data from the target domain. Our results also show that *do*-calculus is complete for *z*-transportability.

Causal effects identifiability (Galles and Pearl, 1995; Tian, 2004; Tian and Pearl, 2002; Shpitser and Pearl, 2006a,b), transportability (Pearl and Bareinboim, 2011; Bareinboim and Pearl, 2012b), *z*-identifiability (Bareinboim and Pearl, 2012a), meta-transportability (Bareinboim and Pearl, 2013b; Lee and Honavar, 2013) and *z*-transportability (introduced in this paper and in Bareinboim and Pearl, 2013a) are all special cases of *meta-identifiability* (Pearl, 2012) which has to do with nonparametric identification of causal effects given *multiple* domains and *arbitrary* information from each domain. Our results suggest several additional special cases of meta-identifiability to consider, including in particular: a generalization of *z*-transportability that allows causal information from possibly different experiments in *multiple* source domains to be combined to facilitate the estimation of a causal effect in a target domain; variants of *z*-identifiability that incorporate constraints on simultaneous controllability of combinations of variables; and combinations thereof.

### Acknowledgments

The authors are grateful to UAI 2013 anonymous reviewers for their thorough reviews. The work of Vasant Honavar while working at the National Science Foundation was supported by the National Science Foundation. Any opinion, finding, and conclusions contained in this article are those of the authors and do not necessarily reflect the views of the National Science Foundation.